\documentclass[a4paper]{article}
\usepackage[T1]{fontenc}
\usepackage{INTERSPEECH2022}
\usepackage{multirow}
\title{Multi-level Fusion of Wav2vec 2.0 and BERT for Multimodal Emotion Recognition}
\name{Zihan Zhao$^1$, Yanfeng Wang$^1{^,}{^2}{^*}$, Yu Wang$^1{^,}{^2}{^*}$}
\address{
  $^1$Cooperative Medianet Innovation Center, Shanghai Jiao Tong University\enspace$^2$Shanghai AI Laboratory
  }
\email{\{zihanzhao, wangyanfeng, yuwangsjtu\}@sjtu.edu.cn}

\begin{document}

\maketitle
\begin{abstract}
The research and applications of multimodal emotion recognition have become increasingly popular recently. However, multimodal emotion recognition faces the challenge of lack of data. To solve this problem, we propose to use transfer learning which leverages state-of-the-art pre-trained models including wav2vec~2.0 and BERT for this task. Multi-level fusion approaches including coattention-based early fusion and late fusion with the models trained on both embeddings are explored. Also, a multi-granularity framework which extracts not only frame-level speech embeddings but also segment-level embeddings including phone, syllable and word-level speech embeddings is proposed to further boost the performance. By combining our coattention-based early fusion model and late fusion model with the multi-granularity feature extraction framework, we obtain result that outperforms best baseline approaches by 1.3\% unweighted accuracy (UA) on the IEMOCAP dataset.

\renewcommand{\thefootnote}{}
\footnotetext{
* Corresponding authors}
\renewcommand{\thefootnote}{}
\footnotetext{
This work was supported by National Natural Science Foundation of China under Grant No. 62106140, Shanghai Science and Technology Committee under Grant No. 21511101100 and Shanghai Key Lab of Digital Media Processing and Transmission (No. 18DZ2270700)}

\end{abstract}
\noindent\textbf{Index Terms}: multimodal emotion recognition, multi-granularity framework, transfer learning

\section{Introduction}

People are highly rich in emotions and expressions of emotions can be found throughout lives. People also possess a strong ability to recognize these emotions in order to generate appropriate responses. Similarly, machines that can recognize emotions to make them more human-like can have many application scenarios \cite{zhou2018emotional,polignano2021towards,ayata2020emotion}. People use multimodal information when perceiving emotions, like text, speech, vision and motion. In this paper, we focus our research on text and speech modalities.


Text is a highly relevant modality for emotion recognition, because the meanings of words and their relations express one's emotion \cite{alswaidan2020survey}. The main challenge of using emotion datasets directly for text emotion recognition is the lack of large-scale data \cite{pepino21_interspeech}. 
This is mainly due to the difficulty posed by the subjective nature of labeling the emotion. One common solution to this problem is to leverage transfer learning-based approaches. For natural language processing, recently proposed pre-trained models such as Bidirectional Encoder Representations from Transformers (BERT) \cite{devlin2018bert}, Generative Pre-training 2 (GPT-2) \cite{radford2019language}, Text-to-Text Transfer Transformer (T5) \cite{raffel2019exploring} become popular because of their excellent performance on a number of important tasks. The difference between GPT-2, T5 and BERT is that GPT-2 and T5 are better at sequence generation while BERT is more appropriate for extracting embeddings \cite{raffel2019exploring,miller2019leveraging,klein2019learning}. For text-based emotion recognition, it is more vital to extract embeddings considering the characteristics of the task, so there are some previous works leveraging BERT for text emotion recognition and achieving promising results \cite{acheampong2021transformer,adoma2020comparative}.

In addition to text, speech is also commonly recognized as a modality with high importance for emotion recognition. When comparing to text, information conveyed by speech such as the intonation and pitch can be used to recognize emotions. Therefore the use of speech for emotion recognition has also been popular, but it faces the similar problem of insufficient data. Similarly, this problem can be mitigated through the use of speech-based pre-trained models such as wav2vec \cite{schneider2019wav2vec}, VQ-wav2vec \cite{baevski2019vq}, wav2vec~2.0 \cite{baevski2020wav2vec}. Recently, wav2vec 2.0 is becoming popular on speech processing tasks such as ASR \cite{zhang2020pushing}, speaker verification \cite{fan2020exploring} and it has also shown promising performance in speech emotion recognition in \cite{pepino21_interspeech}.

Given the importance of both text and speech for emotion recognition tasks, it is natural to leverage multimodal models that combine text and speech information to improve the performance of the emotion recognition and there are a number of works have explored this \cite{liu20b_interspeech,makiuchi2021multimodal,chen20b_interspeech,n20_interspeech,kumar21d_interspeech,LIU20221}. However, to the best of our knowledge, few of them have yet explored how to effectively combine pre-trained models for multimodal emotion recognition. The recent work in~\cite{makiuchi2021multimodal} considers both wav2vec 2.0 and BERT for emotion recognition. However, it mainly focuses on disentanglement representation learning but not on multimodal fusion, and the only fusion method considered is the score fusion. In this paper, we explore extensively the fusion of wav2vec 2.0 and BERT-based embeddings and models for the multimodal emotion recognition. In other recent papers with state-of-the-art results, \cite{kumar21d_interspeech} utilizes attention-based Gated Recurrent Unit for speech and BERT for text respectively and then concatenates them together to get the final prediction. Furthermore, \cite{LIU20221} proposes a method combining self-attentional bidirectional contextual LSTM and self-attentional multi-channel CNN with multi-scale fusion framework including feature-level fusion and decision-level fusion. We summarize our major contributions as follows:
\begin{itemize}
\item We explore the multi-level fusion of text and speech for  emotion recognition leveraging both wav2vec 2.0 and BERT with both early fusion and late fusion, and also the combination of them. For the early fusion, both simple embedding concatenation and more sophisticated attention mechanism-based fusion methods~\cite{han2021bi, siriwardhana2020jointly} are investigated.  To the best of our knowledge, this is the first work that explores the multi-level fusion of state-of-the-art pre-trained model for the multimodal emotion recognition task.
\item We propose a novel multi-granularity fusion framework which makes use of not only frame-level speech embeddings but also explores segment-level speech embeddings including word, syllable and phone-level embeddings to further improve the recognition performance.
\item The proposed models are experimented on the popular IEMOCAP dataset~\cite{busso2008iemocap}. Experimental results show that they can outperform existing state-of-the-art multimodal approaches using both speech and text modalities.
\end{itemize}

\section{Proposed Methods}
For emotion recognition, we mainly explore two fusion mechanisms including  late fusion and early fusion, which can also be combined to further boost the performance. In all three models we utilize multi-granularity frameworks to extract embeddings. It is worth noting that in this part we discuss the scenario with all the considered segment-level embeddings, while in the experiments we will explore the performance of both single and combination of these embeddings. 

\subsection{Multi-granularity Framework}
In this paper, we explore a multi-granularity framework to extract speech embeddings from multiple levels of subwords. As illustrated in the blue part of Figure \ref{1}, the embeddings wav2vec~2.0 extracted are normally frame-level embeddings and it has shown to be effective in obtaining abundant frame-level information. However, it lacks the ability to capture segment-level information which is useful for emotion recognition. Thus, in addition to frame-level embeddings, we introduce segment-level embeddings including word, phone and syllable-level embeddings which are closely related to the prosody \cite{arciuli2007and,du2021mixture,alex2020attention}. Prosody can convey characteristics of the utterance like emotional state \cite{wu2010emotion,rao2013emotion} because it contains the information of the cadence of speech signals. As a result, segment-level embeddings may be helpful for multimodal emotion recognition. 

Using force alignment approaches, the temporal boundaries of phonemes can be obtained, which can then be grouped to get the boundaries of the syllables. Force alignment information is provided in \cite{busso2008iemocap}. The speech segments corresponding to those units can be extracted thereafter. The segment-level embeddings can then be obtained by:
\begin{equation}
\textbf{u}_k^{i,l,n} = \frac{{\sum\nolimits_{f = s_k^{i,l,n}}^{e_k^{i,l,n}} {\textbf{u}_f^{F,l,n}} }}{{e_k^{i.l,n} - s_k^{i,l,n}}}, i \in \rm{\{P,W,S\}}
\nonumber
\end{equation}
Here ${\rm{\textbf{u}}}_k^{i,l,n}$ is the $k$\textsuperscript{th} segmentation of the segment-level embedding of the $l$\textsuperscript{th} layer of the $n$\textsuperscript{th} sample, $f$ is the $f$\textsuperscript{th} frame of the frame-level embedding, $s$ is the starting frame, $e$ is the end frame and $\rm{F}$, $\rm{P}$, $\rm{W}$, $\rm{S}$ represent frame, phone, word and syllable-level speech embeddings respectively. Thereafter, we apply weighted average for layers, following \cite{pepino21_interspeech}, where they took embeddings from all 12 transformer encoder layers in wav2vec~2.0 as the frame-level embeddings and then learned a weight $w_l$ for each of the layer. Then, the weighted embeddings are obtained by:
\begin{equation}
\begin{split}
{\textbf{u}}_k^{i,n} = {\rm{LN}}\left( {\frac{{\sum\nolimits_l {w_l^i{\textbf{u}}_k^{i,l,n}} }}{{\sum\nolimits_l {w_l^i} }}} \right),i \in \{ {\rm{P}},{\rm{W}},{\rm{S}},{\rm{F}},{\rm{T}}\} 
\label{U}
\end{split}
\end{equation}
A similar approach is applied for extracting text embeddings. Here $\rm{LN}$ represents Layer Normalization \cite{ba2016layer}, $w_l^i$ is the learnable weight for the layer $l$ and $\rm{T}$ represents text embeddings.

\subsection{Late Fusion}
\begin{figure}[t]
  \centering
  \includegraphics[width=\linewidth]{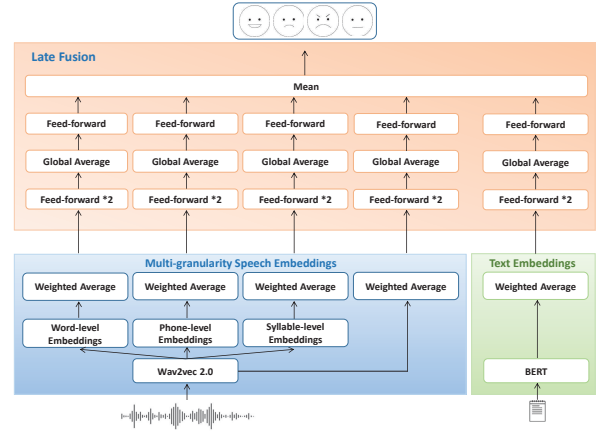}
  \caption{Proposed late fusion model.}
  \label{1}
  
\end{figure}
The orange part of Figure \ref{1} illustrates the late fusion model. There are five branches in the model and for every branch of the late fusion model we use the same structure as that used in \cite{pepino21_interspeech}. First, the extracted embeddings $\textbf{u}_k^{i,n}$ obtained using Equation \ref{U} are input into two feed-forward layers:
\begin{equation}
{\rm{\textbf{h}}}_k^{i,n}{{ = }}{\rm{Relu}}({{\textbf{W}_1^{i}}}\textbf{u}_k^{i,n} + {b_1^{i}}),i \in \rm{\{ P,W,S,F,T\}}
\nonumber
\end{equation}
\begin{equation}
\begin{split}
\textbf{l}_k^{i,n} = {\rm{Relu}}({\textbf{W}_2^{i}}{\rm{\textbf{h}}}_k^{i,n} + {b_2^{i}}),
i \in \rm{\{ P,W,S,F,T\}}
\nonumber
\end{split}
\end{equation}

Then a global average module is applied to fuse embeddings from different segmentation or frames or wordpieces:
\begin{equation}
{\textbf{l}^{i,n}} = \frac{{\sum\nolimits_k {\textbf{l}_k^{i,n}} }}{{{l^{i,n}}}}
\label{ga}
\end{equation}
Here $l^{i,n}$ is the sequence length. Finally, the embeddings are sent to the last feed-forward layer to generate logits, and logits from different branches are added together to generate prediction ${\tilde {\textbf{y}}}^n$ for the late fusion model:
\begin{equation}
{{\tilde {\textbf{y}}}^n} = {\rm{softmax}}\left( {\sum\limits_i {\left( {{{\textbf{W}}_3^{i}}{{\textbf{l}}^{i,n}} + {b_3^{i}}} \right)} } \right)
\label{latefusionequation}
\end{equation}
\subsection{Early Fusion}
\begin{figure}[t]
  \centering
  \includegraphics[width=\linewidth]{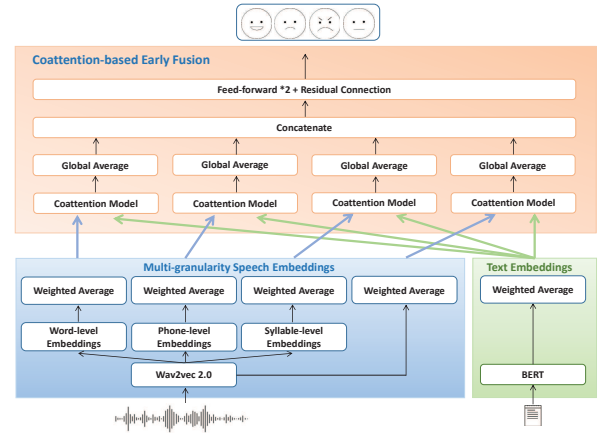}
  \caption{Proposed coattention-based early fusion model.}
  \label{3}
  \vspace{-0.5cm}
\end{figure}
 Figure \ref{3} illustrates the coattention-based early fusion model we adopt. Here multiple levels including phone, word, syllable and frame-level speech embeddings are sent to the coattention model with text embeddings respectively:

\begin{equation}
\begin{split}
{\textbf{c}^{i,n}} = {\rm{GA}}\left( {{\textbf{U}^{\rm{T},n}} \otimes {\textbf{U}^{i,n}}} \right),i \in \rm{\{ P,W,S,F\}}
\nonumber
\end{split}
\end{equation}
where the coattention operation is denoted as $\otimes$, and $\rm{GA}$ represents the global average as the same in Equation \ref{ga}, ${\textbf{U}^{i,n}} = \left[ {\textbf{u}_1^{i,n},\textbf{u}_2^{i,n},...,\textbf{u}_{{l^{i,n}}}^{i,n}} \right]$ and is of size $[seq\_length,embed\_size]$, where $\textbf{u}_k^{i,n}$ is calculated using Equation \ref{U}. $\textbf{c}^{i,n}$ is the embedding of the $n$\textsuperscript{th} sample produced by the coattention model and the global average module, the size of which is $[embed\_size]$. Then the embeddings are concatenated together as:
\begin{equation}
{{\bf{c}}^n} = {{\bf{c}}^{{\rm{P}},n}} \oplus {{\bf{c}}^{{\rm{W}},n}} \oplus {{\bf{c}}^{{\rm{S}},n}} \oplus {{\bf{c}}^{{\rm{F}},n}}
\nonumber
\end{equation}
where the concatenation operation is denoted as $\oplus$. Finally, the concatenated embedding is sent to feed-forward layers for multi-granularity fusion and outputs the estimated posterior probability of the emotion classes ${{\tilde {\textbf{y}}}^n}$:
\begin{equation}
{{\textbf{g}}^n} = {\rm{LN}}\left( {{\rm{Relu}}\left( {{{\textbf{W}}_4}{{\textbf{c}}^n} + {b_4}} \right) + {{\textbf{c}}^n}} \right)
\nonumber
\end{equation}
\begin{equation}
{{\tilde {\textbf{y}}}^n} = {\rm{softmax}}\left( {{{\textbf{W}}_5}{{\textbf{g}}^n} + {b_5}} \right)
\label{earlyfusioneqation}
\end{equation}

\begin{figure}[t]
  \centering
  \includegraphics[width=\linewidth]{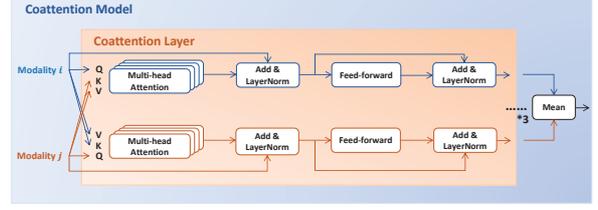}
  \caption{Coattention model.}
  \label{2}
  \vspace{-0.3cm}
\end{figure}

 The structure of the coattention model in Figure \ref{3} is illustrated in Figure \ref{2}. There are two branches in one coattention layer, each of which has the same structure but different modalities as Q or K,V \cite{lu2019vilbert}. Without loss of generality, the calculation process of one-head attention is given here \cite{vaswani2017attention}:
\begin{equation}
{{\textbf{Q}}}^n = {\textbf{W}_6^{i}}\textbf{U}^{i,n} + {b_6^{i}}
\nonumber
\end{equation}
\begin{equation}
\textbf{K}^n = {\textbf{W}_7^{j}}\textbf{U}^{j,n} + {b_7^{j}}
\nonumber
\end{equation}
\begin{equation}
\textbf{V}^n = {\textbf{W}_8^{j}}\textbf{U}^{j,n} + {b_8^{j}}
\nonumber
\end{equation}
where the notations $i$ and $j$ indicate that they are from different modalities. Then we can get the processed embeddings as:
\begin{equation}
{\rm{Attention}}\left( {{{\textbf{Q}}^n},{{\textbf{K}}^n},{{\textbf{V}}^n}} \right) = {\rm{softmax}}\left( {\frac{{{{\textbf{Q}}^n}{{\textbf{K}}^n}^ \top }}{{\sqrt d }}} \right){{\textbf{V}}^n}
\nonumber
\end{equation}
where $d$ represents the dimension of the embeddings. Multi-head attention performs this process multiple times in order to learn information from various representation subspaces. After this, the model follows the classic transformer encoder structure in every branch. In our configuration, the coattention layer is repeated three times. Therefore, the output from every branch can incorporate enough information from another modality and as a result they are almost identical in our experiment, so we  simply calculate the mean to obtain the final output for our coattention model in order to reduce the consumption of computational resources.

 Concatenation-based early fusion is also investigated for comparison. For this fusion approach, we concatenate text and frame-level speech embeddings before feeding them into feed-forward layers which is the same as one branch of our late fusion model.

\subsection{The Combination of Early Fusion and Late Fusion}
We have introduced the models and processes for the early fusion and late fusion approaches in the previous two subsections. In this subsection, we will describe the combination of them to achieve better performance. The reason for the combination is that early fusion merges two modalities with low-level pre-trained features while late fusion merges them with high-level features. Thus, these two fusion schemes make predictions based on embeddings from various levels and it is expected that they can provide complementary information that can be leveraged to boost the performance.

In this work, score combination of best early-fused and late-fused models is considered.  Their output logits are averaged as the combined prediction.

\section{Experiments and Discussion}
\subsection{Datasets}
The IEMOCAP dataset \cite{busso2008iemocap} is an acted, multimodal and multispeaker dataset with five sessions. Following the work in \cite{pepino21_interspeech}, we use 4 emotional classes: anger, happiness, sadness and neutral. We relabeled excitement utterances as happiness and discard utterances from other classes, which is a common practice using IEMOCAP. As a result, the number of utterances representing angry, happy, sad and neutral were 1103, 1636, 1084 and 1708 respectively. A 5-fold cross-validation (CV) configuration was implemented to evaluate our model, leaving one session out as test set in each fold, since it is the regular evaluation method for IEMOCAP. It is worth mentioning that there was no fixed manner to determine the validation set \cite{DBLP:journals/corr/abs-1802-05630}, so we randomly selected 10\% of the utterances in the training set as the validation set in each fold.

\subsection{Settings and Metrics}
 Cross-entropy loss was adopted as our loss function, and an Adam optimizer~\cite{kingma2014adam} was applied using a learning rate of 1e-3 for the feed-forward layer-based models and 5e-5 for transformer-based models. The models were trained using a batch size of 32 and early stopping was also applied. Dropout~\cite{srivastava2014dropout} was applied with a probability of 0.2 after every feed-forward layer except the output layer to prevent overfitting. We use wav2vec~2.0-base and BERT-base un-cased models, both of which have 768-dimensional embeddings. All speech samples are normalized by global normalization which is a frequently used setting for this dataset.

We adopted the widely used metric unweighted accuracy (UA) as our evaluation metric. The results were taken from an average of 25 experiments. (we implemented five-fold cross-validation for 5 times. )

\begin{table}[]
\caption{Unweighted accuracy (UA) of the 5-fold CV results using single modality.}
\label{single}
\centering
\begin{tabular}{c|c|c}
\toprule
\textbf{Embeddings} & \textbf{Linear} & \textbf{Transformer Encoder} \\ \midrule
Text                & \textbf{68.12}\% & \textbf{67.29}\%              \\ 
Frame               & 65.44\%          & 62.85\%                       \\ 
Phone               & 64.24\%          & 62.79\%                       \\ 
Syllable            & 63.86\%          & 63.26\%                       \\ 
Word                & 64.48\%          & 62.38\%                       \\ \bottomrule
\end{tabular}

\end{table}

\subsection{Results of the Single Modality Models}
Table \ref{single} shows the results of the single modality. "Linear" means the linear model used in \cite{pepino21_interspeech}, which is also adopted in every branch of our late fusion model. The "Transformer Encoder" has three layers of transformer encoders followed by two feed-forward layers, which is used for comparison with our coattention-based early fusion model. We can observe that linear models outperform transformer encoder models in every embeddings. This is consistent with the observation in~\cite{siriwardhana2020jointly}, and we think it may be caused by the overfitting problem. It also can be seen that the text embeddings always outperform speech embeddings. This is because there are not many complex contexts where speech is more effective, such as sarcasm, in this dataset. When comparing to the three segment-level speech embeddings, frame-level speech embeddings perform better in linear models, while syllable-level embeddings perform better in transformer encoder models. The segment-level speech embeddings have more information linked to prosody, thus it can provide more information when combined with the frame-level speech embeddings and text embeddings. This will be validated in the next subsection. It is worth noting that when using frame-level speech embeddings, it corresponds to a reproduction of the approach in~\cite{pepino21_interspeech} and the UA (65.44\%) is very similar to that (65.80\%) given in~\cite{pepino21_interspeech}.

\begin{table}[]
\caption{Unweighted accuracy (UA) of the 5-fold CV results of coattention-based early fusion and late fusion models with different input embeddings. F - frame, P - phone, S - syllable, W - word, Coattention - coattention-based early fusion}
\label{main_result}
\centering
\resizebox{\linewidth}{!}{
\begin{tabular}{cc|cc}
\toprule
\multicolumn{2}{c|}{\textbf{Embeddings}}                     & \multicolumn{2}{c}{\textbf{Fusion}}                              \\ \midrule
\multicolumn{1}{c|}{\textbf{Text}}         & \textbf{Speech} & \multicolumn{1}{c|}{\textbf{Coattention}} & \textbf{Late Fusion} \\ \midrule
\multicolumn{1}{c|}{\multirow{8}{*}{BERT}} & F               & \multicolumn{1}{c|}{74.28\%}               & 74.88\%               \\ \cline{2-4} 
\multicolumn{1}{c|}{}                      & F+P             & \multicolumn{1}{c|}{\textbf{74.57}\%}      & 75.15\%               \\
\multicolumn{1}{c|}{}                      & F+S             & \multicolumn{1}{c|}{74.05\%}               & \textbf{75.80}\%      \\
\multicolumn{1}{c|}{}                      & F+W             & \multicolumn{1}{c|}{74.27\%}               & 75.22\%               \\ \cline{2-4} 
\multicolumn{1}{c|}{}                      & F+P+S           & \multicolumn{1}{c|}{73.35\%}               & 74.67\%               \\
\multicolumn{1}{c|}{}                      & F+P+W           & \multicolumn{1}{c|}{73.96\%}               & 75.52\%               \\
\multicolumn{1}{c|}{}                      & F+S+W           & \multicolumn{1}{c|}{73.53\%}               & 75.59\%               \\ \cline{2-4} 
\multicolumn{1}{c|}{}                      & F+P+S+W         & \multicolumn{1}{c|}{73.64\%}               & 74.60\%       \\
\bottomrule
\end{tabular}}
\vspace{-0.5cm}
\end{table}

\subsection{Results of the Multi-level Fusions}

Table~\ref{main_result} shows the results of our coattention-based early fusion model using Equation~\ref{earlyfusioneqation} and late fusion model using Equation~\ref{latefusionequation} with combinations of different segment-level speech embeddings. It shows that, compared to unimodal results in table~\ref{single}, any combination can yield substantial performance improvement. It demonstrates that there is indeed complementarity between two modalities on this task. It can also be seen from Table~\ref{main_result} that late fusion models generally have better results than early fusion models. The multi-granularity frame works quite well on the late fusion model and also achieves better results in the early fusion model than inputs without segment-level speech embeddings. It can be seen from both fusions that adding more embeddings cannot always lead to better performance, thus the improvement brought by the introduction of segment-level speech embeddings does not result from the effect of ensemble learning, but result from the introduction of prosodic information which is relevant to emotion recognition.

\begin{table}[]
\caption{Performance comparison between our models and
state-of-the-art multimodal models on the IEMOCAP dataset. F - frame, P - phone, S - syllable, W - word, Concatenation - concatenation-based early fusion, Coattention - coattention-based early fusion}
\label{sota}
\centering
\resizebox{\linewidth}{!}{
\begin{tabular}{ccc|c}
\toprule
\multicolumn{3}{c|}{\textbf{Proposed Methods}}                                                        & \multirow{2}{*}{\textbf{UA}} \\ \cline{1-3}
\multicolumn{2}{c|}{\textbf{Embeddings}}                                & \textbf{Fusion}             &                              \\ \midrule
\multicolumn{1}{c|}{\multirow{4}{*}{BERT}} & \multicolumn{1}{c|}{F+P}     & Concatenation               & 71.29\%                       \\
\multicolumn{1}{c|}{}                      & \multicolumn{1}{c|}{F+P}     & Coattention                 & 74.57\%                       \\

\multicolumn{1}{c|}{}                      & \multicolumn{1}{c|}{F+S}   & Late Fusion                 & 75.80\%                       \\
\multicolumn{1}{c|}{}                      & \multicolumn{1}{c|}{F+P+S} & Coattention and Late Fusion & \textbf{76.31}\%              \\ \toprule
\multicolumn{3}{c|}{\textbf{Baseline Methods}}                                                        & \textbf{UA}                  \\ \midrule
\multicolumn{3}{l|}{GBAN (Liu et al. \cite{liu20b_interspeech}, 2020)}                                                                                & \multicolumn{1}{l}{70.08\%}   \\
\multicolumn{3}{l|}{STSER (Chen et al.\cite{chen20b_interspeech}, 2020)}                                                                                & \multicolumn{1}{l}{72.05\%}   \\
\multicolumn{3}{l|}{Krishna et al. \cite{n20_interspeech}, 2020}                                                                                & \multicolumn{1}{l}{72.82\%}   \\
\multicolumn{3}{l|}{Makiuchi et al. \cite{makiuchi2021multimodal}, 2021}                                                                                & \multicolumn{1}{l}{73.00\%}   \\
\multicolumn{3}{l|}{Kumar et al. \cite{kumar21d_interspeech}, 2021}                                                                                & \multicolumn{1}{l}{75.00\%}   \\
\multicolumn{3}{l|}{Liu et al. \cite{LIU20221}, 2022}                                                                                & \multicolumn{1}{l}{75.05\%}   \\ \bottomrule
\end{tabular}}
\vspace{-0.5cm}
\end{table}

\subsection{Comparison of State-of-the-art Approaches and the Proposed Approaches}


This subsection further validates the effectiveness of the proposed approaches. In the first block of Table~\ref{sota}, the effectiveness of the coattention-based early fusion approach is first validated, as it can be seen that with the frame-level and phone-level speech embeddings, the coattention can give good performance gains when comparing with the concatenation-based fusion, improving the UA by about 3\%. Here we provide the comparison between the coattention and the concatenation-based early fusion only with frame-level and phone-level speech embeddings due to space limitation, but there are similar results in the remaining cases. We copy the best results of coattention-based early fusion (74.57\%) and late fusion (75.80\%) from Table \ref{main_result}. This demonstrates that by combining the best configurations of both early (F+P) and late (F+S) fusions (so used embeddings are F+P+S), it gives another 0.5\% UA increase (76.31\%) over the best performing late fusion model. This proves that combining the multi-modal models at multiple levels can improve the performance.

In the second block of Table~\ref{sota}, the proposed approaches are also compared with other state-of-the-art multi-modal speech-text emotion recognition approaches. For a fair comparison, all the approaches are based on the 5-fold cross validation configuration and use the same way of preprocessing of IEMOCAP. It shows that our model using text and only frame-level speech embeddings already surpasses the performance of most baseline approaches, demonstrating the advantage of combining wav2vec~2.0 and BERT. In addition, the best multi-level fusion surpasses the best baseline approach by 1.3\% UA.

\section{Conclusion and Future Work}
In this paper, we have proposed to leverage the state-of-the-art wav2vec~2.0 and BERT embeddings with a multi-level fusion framework to mitigate the issue of data sparsity in multimodal emotion recognition and also explored the multi-granularity framework. Our best fusion configuration achieves an accuracy of 76.31\% UA for the 5-fold CV on IEMOCAP. In the future, we plan to explore the fine-tuning of wav2vec 2.0 and BERT in our model while in this paper we fix them and only use them as feature extractor.


\bibliographystyle{IEEEtran}

\bibliography{mybib}

\begin{thebibliography}{10}
\providecommand{\url}[1]{#1}
\csname url@samestyle\endcsname
\providecommand{\newblock}{\relax}
\providecommand{\bibinfo}[2]{#2}
\providecommand{\BIBentrySTDinterwordspacing}{\spaceskip=0pt\relax}
\providecommand{\BIBentryALTinterwordstretchfactor}{4}
\providecommand{\BIBentryALTinterwordspacing}{\spaceskip=\fontdimen2\font plus
\BIBentryALTinterwordstretchfactor\fontdimen3\font minus
  \fontdimen4\font\relax}
\providecommand{\BIBforeignlanguage}[2]{{%
\expandafter\ifx\csname l@#1\endcsname\relax
\typeout{** WARNING: IEEEtran.bst: No hyphenation pattern has been}%
\typeout{** loaded for the language `#1'. Using the pattern for}%
\typeout{** the default language instead.}%
\else
\language=\csname l@#1\endcsname
\fi
#2}}
\providecommand{\BIBdecl}{\relax}
\BIBdecl

\bibitem{zhou2018emotional}
H.~Zhou, M.~Huang, T.~Zhang, X.~Zhu, and B.~Liu, ``Emotional chatting machine:
  Emotional conversation generation with internal and external memory,'' in
  \emph{Proceedings of the AAAI Conference on Artificial Intelligence},
  vol.~32, no.~1, 2018.

\bibitem{polignano2021towards}
M.~Polignano, F.~Narducci, M.~de~Gemmis, and G.~Semeraro, ``Towards
  emotion-aware recommender systems: an affective coherence model based on
  emotion-driven behaviors,'' \emph{Expert Systems with Applications}, vol.
  170, p. 114382, 2021.

\bibitem{ayata2020emotion}
D.~Ayata, Y.~Yaslan, and M.~E. Kamasak, ``Emotion recognition from multimodal
  physiological signals for emotion aware healthcare systems,'' \emph{Journal
  of Medical and Biological Engineering}, vol.~40, no.~2, pp. 149--157, 2020.

\bibitem{alswaidan2020survey}
N.~Alswaidan and M.~E.~B. Menai, ``A survey of state-of-the-art approaches for
  emotion recognition in text,'' \emph{Knowledge and Information Systems},
  vol.~62, no.~8, pp. 2937--2987, 2020.

\bibitem{pepino21_interspeech}
L.~Pepino, P.~Riera, and L.~Ferrer, ``{Emotion Recognition from Speech Using
  Wav2vec 2.0 Embeddings},'' in \emph{Proc. Interspeech}, 2021, pp. 3400--3404.

\bibitem{devlin2018bert}
J.~Devlin, M.-W. Chang, K.~Lee, and K.~Toutanova, ``{BERT}: Pre-training of
  deep bidirectional transformers for language understanding,'' \emph{arXiv
  preprint arXiv:1810.04805}, 2018.

\bibitem{radford2019language}
A.~Radford, J.~Wu, R.~Child, D.~Luan, D.~Amodei, I.~Sutskever \emph{et~al.},
  ``Language models are unsupervised multitask learners,'' \emph{OpenAI blog},
  vol.~1, no.~8, p.~9, 2019.

\bibitem{raffel2019exploring}
C.~Raffel, N.~Shazeer, A.~Roberts, K.~Lee, S.~Narang, M.~Matena, Y.~Zhou,
  W.~Li, and P.~J. Liu, ``Exploring the limits of transfer learning with a
  unified text-to-text transformer,'' \emph{arXiv preprint arXiv:1910.10683},
  2019.

\bibitem{miller2019leveraging}
D.~Miller, ``Leveraging {BERT} for extractive text summarization on lectures,''
  \emph{arXiv preprint arXiv:1906.04165}, 2019.

\bibitem{klein2019learning}
T.~Klein and M.~Nabi, ``Learning to answer by learning to ask: Getting the best
  of {GPT}-2 and {BERT} worlds,'' \emph{arXiv preprint arXiv:1911.02365}, 2019.

\bibitem{acheampong2021transformer}
F.~A. Acheampong, H.~Nunoo-Mensah, and W.~Chen, ``Transformer models for
  text-based emotion detection: a review of {BERT}-based approaches,''
  \emph{Artificial Intelligence Review}, vol.~54, no.~8, pp. 5789--5829, 2021.

\bibitem{adoma2020comparative}
A.~F. Adoma, N.-M. Henry, and W.~Chen, ``Comparative analyses of {BERT,
  RoBERTa, DistilBERT, and XLNet} for text-based emotion recognition,'' in
  \emph{17th International Computer Conference on Wavelet Active Media
  Technology and Information Processing (ICCWAMTIP)}.\hskip 1em plus 0.5em
  minus 0.4em\relax IEEE, 2020, pp. 117--121.

\bibitem{schneider2019wav2vec}
S.~Schneider, A.~Baevski, R.~Collobert, and M.~Auli, ``wav2vec: Unsupervised
  pre-training for speech recognition,'' \emph{arXiv preprint
  arXiv:1904.05862}, 2019.

\bibitem{baevski2019vq}
A.~Baevski, S.~Schneider, and M.~Auli, ``vq-wav2vec: Self-supervised learning
  of discrete speech representations,'' \emph{arXiv preprint arXiv:1910.05453},
  2019.

\bibitem{baevski2020wav2vec}
A.~Baevski, Y.~Zhou, A.~Mohamed, and M.~Auli, ``wav2vec 2.0: A framework for
  self-supervised learning of speech representations,'' \emph{Advances in
  Neural Information Processing Systems}, vol.~33, pp. 12\,449--12\,460, 2020.

\bibitem{zhang2020pushing}
Y.~Zhang, J.~Qin, D.~S. Park, W.~Han, C.-C. Chiu, R.~Pang, Q.~V. Le, and Y.~Wu,
  ``Pushing the limits of semi-supervised learning for automatic speech
  recognition,'' \emph{arXiv preprint arXiv:2010.10504}, 2020.

\bibitem{fan2020exploring}
Z.~Fan, M.~Li, S.~Zhou, and B.~Xu, ``Exploring wav2vec 2.0 on speaker
  verification and language identification,'' \emph{arXiv preprint
  arXiv:2012.06185}, 2020.

\bibitem{liu20b_interspeech}
P.~Liu, K.~Li, and H.~Meng, ``{Group Gated Fusion on Attention-Based
  Bidirectional Alignment for Multimodal Emotion Recognition},'' in \emph{Proc.
  Interspeech}, 2020, pp. 379--383.

\bibitem{makiuchi2021multimodal}
M.~R. Makiuchi, K.~Uto, and K.~Shinoda, ``{Multimodal Emotion Recognition with
  High-level Speech and Text Features},'' in \emph{Proc. ASRU}, 2021.

\bibitem{chen20b_interspeech}
M.~Chen and X.~Zhao, ``{A Multi-Scale Fusion Framework for Bimodal Speech
  Emotion Recognition},'' in \emph{Proc. Interspeech}, 2020, pp. 374--378.

\bibitem{n20_interspeech}
K.~D. N. and A.~Patil, ``{Multimodal Emotion Recognition Using Cross-Modal
  Attention and 1D Convolutional Neural Networks},'' in \emph{Proc.
  Interspeech}, 2020, pp. 4243--4247.

\bibitem{kumar21d_interspeech}
P.~Kumar, V.~Kaushik, and B.~Raman, ``{Towards the Explainability of Multimodal
  Speech Emotion Recognition},'' in \emph{Proc. Interspeech}, 2021, pp.
  1748--1752.

\bibitem{LIU20221}
Y.~Liu, H.~Sun, W.~Guan, Y.~Xia, and Z.~Zhao, ``Multi-modal speech emotion
  recognition using self-attention mechanism and multi-scale fusion
  framework,'' \emph{Speech Communication}, vol. 139, pp. 1--9, 2022.

\bibitem{han2021bi}
W.~Han, H.~Chen, A.~Gelbukh, A.~Zadeh, L.-p. Morency, and S.~Poria,
  ``Bi-bimodal modality fusion for correlation-controlled multimodal sentiment
  analysis,'' in \emph{Proceedings of the International Conference on
  Multimodal Interaction}, 2021, pp. 6--15.

\bibitem{siriwardhana2020jointly}
S.~Siriwardhana, A.~Reis, R.~Weerasekera, and S.~Nanayakkara, ``Jointly
  fine-tuning" bert-like" self supervised models to improve multimodal speech
  emotion recognition,'' \emph{arXiv preprint arXiv:2008.06682}, 2020.

\bibitem{busso2008iemocap}
C.~Busso, M.~Bulut, C.-C. Lee, A.~Kazemzadeh, E.~Mower, S.~Kim, J.~N. Chang,
  S.~Lee, and S.~S. Narayanan, ``{IEMOCAP}: Interactive emotional dyadic motion
  capture database,'' \emph{Language resources and evaluation}, vol.~42, no.~4,
  pp. 335--359, 2008.

\bibitem{arciuli2007and}
J.~Arciuli and L.~M. Slowiaczek, ``The where and when of linguistic word-level
  prosody,'' \emph{Neuropsychologia}, vol.~45, no.~11, pp. 2638--2642, 2007.

\bibitem{du2021mixture}
C.~Du and K.~Yu, ``Mixture density network for phone-level prosody modelling in
  speech synthesis,'' \emph{arXiv e-prints}, pp. arXiv--2102, 2021.

\bibitem{alex2020attention}
S.~B. Alex, L.~Mary, and B.~P. Babu, ``Attention and feature selection for
  automatic speech emotion recognition using utterance and syllable-level
  prosodic features,'' \emph{Circuits, Systems, and Signal Processing},
  vol.~39, no.~11, pp. 5681--5709, 2020.

\bibitem{wu2010emotion}
C.-H. Wu and W.-B. Liang, ``Emotion recognition of affective speech based on
  multiple classifiers using acoustic-prosodic information and semantic
  labels,'' \emph{IEEE Transactions on Affective Computing}, vol.~2, no.~1, pp.
  10--21, 2010.

\bibitem{rao2013emotion}
K.~S. Rao, S.~G. Koolagudi, and R.~R. Vempada, ``Emotion recognition from
  speech using global and local prosodic features,'' \emph{International
  journal of speech technology}, vol.~16, no.~2, pp. 143--160, 2013.

\bibitem{ba2016layer}
J.~L. Ba, J.~R. Kiros, and G.~E. Hinton, ``Layer normalization,'' \emph{arXiv
  preprint arXiv:1607.06450}, 2016.

\bibitem{lu2019vilbert}
J.~Lu, D.~Batra, D.~Parikh, and S.~Lee, ``Vilbert: Pretraining task-agnostic
  visiolinguistic representations for vision-and-language tasks,''
  \emph{Advances in neural information processing systems}, vol.~32, 2019.

\bibitem{vaswani2017attention}
A.~Vaswani, N.~Shazeer, N.~Parmar, J.~Uszkoreit, L.~Jones, A.~N. Gomez,
  {\L}.~Kaiser, and I.~Polosukhin, ``Attention is all you need,''
  \emph{Advances in neural information processing systems}, vol.~30, 2017.

\bibitem{DBLP:journals/corr/abs-1802-05630}
C.~Etienne, G.~Fidanza, A.~Petrovskii, L.~Devillers, and B.~Schmauch, ``Speech
  emotion recognition with data augmentation and layer-wise learning rate
  adjustment,'' \emph{CoRR}, vol. abs/1802.05630, 2018.

\bibitem{kingma2014adam}
D.~P. Kingma and J.~Ba, ``Adam: A method for stochastic optimization,''
  \emph{arXiv preprint arXiv:1412.6980}, 2014.

\bibitem{srivastava2014dropout}
N.~Srivastava, G.~Hinton, A.~Krizhevsky, I.~Sutskever, and R.~Salakhutdinov,
  ``Dropout: a simple way to prevent neural networks from overfitting,''
  \emph{The journal of machine learning research}, vol.~15, no.~1, pp.
  1929--1958, 2014.

\end{thebibliography}


\end{document}